\documentclass[letterpaper, 10 pt, conference]{ieeeconf}  

\IEEEoverridecommandlockouts                              
\usepackage{amsmath} 
\usepackage{amssymb}  
\usepackage{mathtools}
\usepackage{multirow}
\usepackage{url}
\usepackage{makecell}
\usepackage{lipsum}  
\usepackage{float}
\usepackage{stfloats}
\usepackage{cite}
\usepackage[colorlinks=true, allcolors=blue]{hyperref}
\usepackage{xcolor}
\usepackage{array}
\usepackage{tabularray}
\usepackage{threeparttable}

\title{\LARGE \bf Hybrid-Neuromorphic Approach for Underwater Robotics Applications: A Conceptual Framework}

\author{ Vidya Sudevan$^{1}$, Fakhreddine Zayer$^{2}$, Sajid Javed$^{1,2}$, Hamad Karki$^{1,3}$, Giulia De Masi$^{1}$, Jorge Dias$^{1,2}$ 
\thanks{$^{1}$Center for Autonomous Robotic Systems, Khalifa University, Abu Dhabi, UAE.}%
\thanks{$^{2}$Khalifa University, Electrical Engineering And Computer Science, Abu Dhabi, UAE.}%
\thanks{$^{3}$Khalifa University, Mechanical and Nuclear Engineering, Abu Dhabi, United Arab Emirates}%
}

\usepackage{graphicx}
\begin{document}
\maketitle

\begin{abstract}
This paper introduces the concept of employing neuromorphic methodologies for task-oriented underwater robotics applications. In contrast to the increasing computational demands of conventional deep learning algorithms, neuromorphic technology, leveraging spiking neural network architectures, promises sophisticated artificial intelligence with significantly reduced computational requirements and power consumption, emulating human brain operational principles. Despite documented neuromorphic technology applications in various robotic domains, its utilization in marine robotics remains largely unexplored. Thus, this article proposes a unified framework for integrating neuromorphic technologies for perception, pose estimation, and haptic-guided conditional control of underwater vehicles, customized to specific user-defined objectives. This conceptual framework stands to revolutionize underwater robotics, enhancing efficiency and autonomy while reducing energy consumption. By enabling greater adaptability and robustness, this advancement could facilitate applications such as underwater exploration, environmental monitoring, and infrastructure maintenance, thereby contributing to significant progress in marine science and technology.
\end{abstract}

\section{Introduction} \label{intro}
\begin{figure}[t]
\centering
\includegraphics[width=0.75\columnwidth]{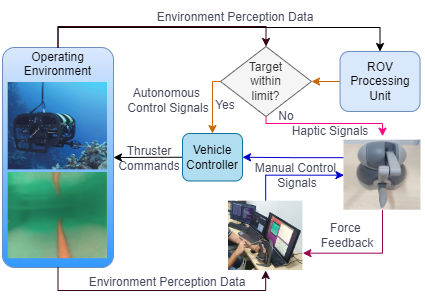}
\vspace{-0.6cm}
\caption{High-level framework of a task-oriented underwater control.}
\label{fig:high_level}
\vspace{-0.6cm}
\end{figure}

In underwater exploration, robotics has emerged as a transformative tool, enabling us to delve into the depths of our oceans with unprecedented precision and efficiency. However, traditional computational approaches in underwater robotics encounter significant hurdles, from high energy consumption to computational complexity, hindering their full potential. As we stand on the brink of a new era in robotics, there arises a pressing need for innovative solutions to propel underwater exploration further. Enter neuromorphic technology—a groundbreaking paradigm inspired by the intricate workings of the human brain. By harnessing the power of spiking neural networks, neuromorphic technology offers a tantalizing promise: the ability to imbue underwater robots with intelligence akin to our own, while simultaneously minimizing computational demands and energy consumption \cite{sharpeshkar2010ultra}.
The field of underwater robotics has witnessed significant advancements in recent years, with robots playing pivotal roles in various applications such as marine exploration, environmental monitoring, and offshore industry operations \cite{di2021monitoring, phillips2016exploring, lu2019treatment}. However, the majority of underwater robotic systems rely on conventional computational approaches, which often entail high computational demands and power consumption \cite{teixeira2020deep, wang2020real, yang2020underwater}. These traditional methods face significant challenges in processing the vast amounts of sensory data collected from underwater environments, including sonar images, video feeds, and environmental sensor data \cite{liu2020underwater, de2021survey}. To illustrate, Fig.~\ref{fig:high_level} provides a high-level representation of a task-oriented underwater control framework, highlighting the complexity involved in integrating various components such as perception, localization, and control. The reliance on conventional computational approaches in such frameworks limits the operational efficiency and autonomy of underwater robots, constraining their ability to perform complex tasks in real-world underwater environments \cite{lanzagorta2019assessing, ejeian2019design}. As a result, there is a growing need for innovative approaches to address these limitations and unlock the full potential of underwater robotics in various domains.

The proposed framework integrates multiple neuromorphic modules into a unified system for task-specific objectives in challenging underwater environments. Aimed at assisting human operators in navigating turbid waters, it focuses on keeping target structures within the operator's visual range. Key modules include Visibility Enhancement and Target Detection, Force Feedback, and Underwater Vehicle Control. Leveraging neuromorphic approaches like spiking-based dehazing and Spiking-YOLO for perception tasks, the framework enhances perception and control. Target pose estimation utilizes a spiking-based rotation localization network, along with Multi-layer Convolutional Spiking Neural Networks and spiking LSTM for vehicle positioning. Offering benefits like adaptability, robustness, and energy efficiency, this framework promises to revolutionize underwater robotics and marine science, with applications spanning exploration, monitoring, and maintenance.

\section{Conceptual Framework of Hybrid- Neuromorphic Approach for Underwater Robotic Applications} \label{sec:overall}

The challenge for both conventional and neuromorphic robots is to design robust and flexible parts to address a wide range of applications, especially when human-robot cooperation is involved. Here, several neuromorphic modules are combined into a unified framework to accomplish task-specific objectives. The diagram illustrating the task-oriented control system for underwater vehicles, presented in Fig.\ref{fig:block_dia}, is designed to assist human operators in navigating through challenging underwater environments characterized by varying levels of turbidity. Its primary purpose is to ensure that the target structure remains within the operator's allowable visual range. The thruster commands are generated based on either Position Based Visual Servoing (PBVS) control or by combined PBVS-Haptic control, depending on the position of the target structure and the acceptable threshold limit. The details of each modules are as follows:

\begin{figure*}[phbt]
\includegraphics[width=0.75\textwidth]{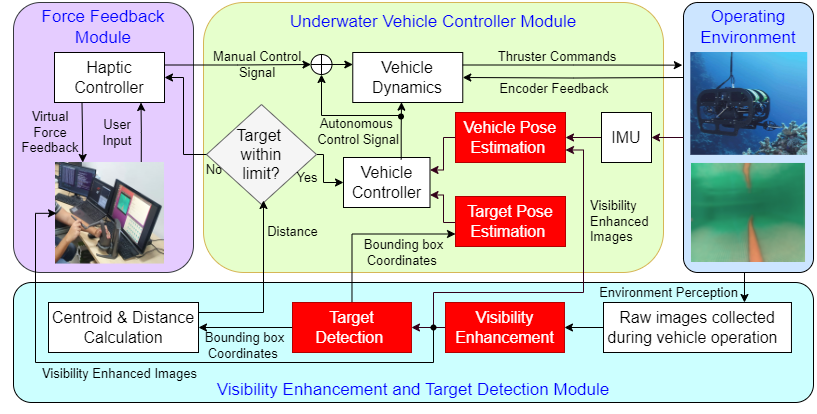}
\centering
\vspace{-0.3cm}
\caption{Schematic representation of the task-driven control framework for underwater vehicles. The highlighted blocks in 'red' denote the incorporation of neuromorphic principles within the depicted modules.}
\label{fig:block_dia}
\end{figure*}
\vspace{-0.1cm}
\subsection{Visibility Enhancement and Target Detection Module}
Underwater environments can have high levels of turbidity, leading to limited visual perception due to strong light absorption and high levels of image distortion caused by backscattered light from suspended particles. The raw images captured are fed to a neuromorphic visibility enhancement module to improve the image quality and aid the target detection process. The bounding box coordinates of the target structure obtained through a neuromorphic target detection module is used to calculate the centroid \(\mathbf{c_p}=\left \{ x_l,y_l \right \}\) of the structure. The centroid pixel values are then used to calculate the haptic rendering force. The details of the neuromorphic visibility enhancement and the target detection module is presented in section.\ref{sec:neuromorphic}.
\vspace{-0.1cm}
\subsection{Force Feedback Module}
The haptic rendering force is calculated only in the lateral direction to ensure that the position of the target structure is within the threshold distance with respect the vehicle. The lateral tracking error is defined as:
\begin{equation}
     e_l(t)=x_c(t)-x_l(t)
    \label{equation:error}
\end{equation}
where \(x_c\) is the central x-coordinate of the image frame. The lateral tracking error \(e_l(t)\) represents the lateral correction that is needed for the vehicle to make the target structure within the threshold limit. Assume that \(\theta_x\) represents the joystick's lateral displacement, which corresponds to the vehicle's translation and yaw rotation. By generating torque in the lateral direction using the DC motor attached to the joystick, the vehicle can be guided toward the desired position, resulting in the lateral tracking error converging to zero. The torque produced around the joystick in the lateral direction is: 
\begin{equation}
    \tau_{j}^{x}=-A\left [ \theta_x(t)-\theta_{x}^{d}(t) \right ]-B\dot{\theta}_x(t)
    \label{equation:torque}
\end{equation}
 The gain constants, A and B, are determined through experimental methods. \(\theta_{x}^{d}\) is the desired value for lateral joystick displacement and is generated using a PD controller to decrease the tracking error and its expression is: 

\begin{equation}
    \theta_{x}^{d}(t)=-\frac{1}{a}\left [ k_p e_l(t)+k_v \dot{e_l(t)} \right ]
    \label{equation:desired_joy}
\end{equation}

The tracking error is monitored continuously and if it exceeds the pre-defined allowable visual range, the torque generated around the joystick axis will \(\tau_{j}^{x}\) or else zero. This implies that if the target structure is within the pre-defined visual level, no haptic feedback will be provided to the human operator.

\begin{equation}
    \tau_{final}=\left\{\begin{matrix}
    \tau_{j}^{x} & ,e_l(t)\geq e_{th}(t)\\ 
    0 & ,e_l(t)< e_{th}(t) 
    \end{matrix}\right.
    \label{equation:locking}
\end{equation}

\subsection{Underwater Vehicle Controller Module}
The forward velocity is held constant where as the vehicle controller module generates the heading correction commands by utilizing the PBVS approach \cite{chaumette2016visual} and will be proportional to \(v_{pbvs}\). The control law \(v_{pbvs}\) based on PBVS is typically derived by exponentially reducing the pose error term \(e_{p}\):
\begin{equation}
    e_{p}=s_{a}-s_{t}=\binom{_{}^{c}\textrm{p}_{a}}{\theta u}-\binom{_{}^{c}\textrm{p}_{t}}{0}=\binom{_{}^{c}\textrm{p}_{a}-_{}^{c}\textrm{p}_{t}}{\theta u}
    \label{equation:pose_error}
\end{equation}
where \(s_{a}\) and \(s_{t}\) are the actual and target 3D pose of the vehicle based on image measurements respectively and is represented by the translation and orientation vector. \(_{}^{c}\textrm{p}_{a}\) is the translation vector of the actual pose of the vehicle and is obtained from the neuromorphic pose estimation module, whereas the \(_{}^{c}\textrm{p}_{t}\) is the translation vector of the target pose. \(\theta u\) gives the axis-angle representation of the rotation matrix. The control law \(v_{pbvs}\) can be derived as:
\begin{equation}
    v_{pbvs}=\hat{\mathbf{L_{s}^{+}}}\dot{e_{p}}=-\lambda \hat{\mathbf{L_{s}^{+}}}e_{p}
    \label{equation:control_law}
\end{equation}
where \(\hat{\mathbf{L_{s}^{+}}}\) is the estimated pseudo-inverse of the interaction matrix \(\mathbf{L_{s}}=\begin{bmatrix}
\mathbf{I_{3}} & -\left [ _{}^{c}\textrm{p}_{a} \right ]_{\times }\\ 
\mathbf{0} & \mathbf{I_{3}} 
\end{bmatrix}\). \(\left [ _{}^{c}\textrm{p}_{a} \right ]_{\times }\) is the skew symmetric matrix representation of the actual translation vector \(_{}^{c}\textrm{p}_{a}\).

The position of the underwater vehicle can drift away from the target position due to unknown disturbances in the underwater environment (the position of the target structure is not within the permissible visual range), resulting in the generation of inaccurate control commands. In such cases, incorporating a conditional controller aids the shared control scheme by activating haptic control alongside the PBVS control scheme, allowing the human operator to re-position the underwater vehicle within the required inspection field. The final control law is represented as:

\begin{equation}
    v_{final}=\left\{\begin{matrix}
    v_{pbvs}+\tau_{final} & ,e_l(t)\geq e_{th}(t)\\ 
    v_{pbvs} & ,e_l(t)< e_{th}(t) 
    \end{matrix}\right.
    \label{equation:final_control_law}
\end{equation}

\section{Neuromorphic Approaches}\label{sec:neuromorphic}
\subsection{Fundamental Blocks}
\subsubsection{\textbf{Spiking Neuron Model}}
The bio-inspired neuromorphic approaches, and in particular spiking neural networks (SNN) may offer increased flexibility for applications in computer vision \cite{meftah2013image}. The literature review reveals that crucial factors in designing a learning algorithm for SNNs include neuron models, synaptic communication, network topology, and information encoding/decoding methods. The analysis of different spiking neuron models for various applications is presented in \cite{koravuna2023exploring}. Leaky Integrate-and-Fire (LIF) \cite{koch1998methods} and Spike Response Model (SRM) \cite{gerstner2014neuronal} are the popular neuron models known for their low computational demands but lack biological plausibility compared to the Hodgkin and Huxley model. A neuron model showcasing the spiking and bursting patterns observed in cortical neurons is introduced in \cite{izhikevich2003simple}. This model merges the biologically accurate Hodgkin-Huxley-type dynamics with the computational effectiveness of integrate-and-fire neurons.  The dynamics of a LIF is given by: 
\begin{equation}
\begin{split}
    \tau_{m}\frac{d V_{m}}{dt}= -V_{m}+\omega *\theta (t-t_{k})
    \label{equation:change_in_membrane_potential}
\end{split}
\end{equation}

It is used to simulate the behavior of a spiking neuron that receives inputs from other neurons via plastic synapses. The LIF model is based on a threshold mechanism and is capable of describing the temporal dynamics of membrane potential (\(V_{m}\)) changes in response to input spikes that are modified by interconnecting synaptic weights. 
The synaptic weight \(\omega\) undergoes modulation due to the incoming spike \(\theta (t-t_{k})\) at time \(t_{k}\). This modulation leads to a current that the post-neuron integrates into its membrane potential. Upon the accumulation of membrane potential to \(V_{th}\),  the LIF neuron resets and emits an output spike to the fan-out synapses.

\subsubsection{\textbf{Data encoding approaches}}
In traditional ANNs, inputs and outputs are analog-valued, while SNNs communicate via spikes or action potentials \cite{gerstner2014neuronal}. Unlike ANNs, SNNs require spike-based encoding schemes for analog input information. The encoding schemes can be broadly classified into (i) rate coding  \cite{guo2021neural}, which involves the number of spikes in a given time period, and (ii) temporal coding \cite{maass2001relevance}, where the precise timing of spikes is considered. Rate coding converts continuous data into a discrete sequence of spikes within a defined time window, utilizing the correlation between firing rates and the intensity of data. The utilization of Time-To-First-Spike (TTFS)  coding was identified as a means to encode information swiftly, particularly for rapid responses within milliseconds, such as tactile stimuli \cite{johansson2004first}, through the utilization of initial spikes.

A more straightforward encoding approach is proposed in \cite{eshraghian2023training} to aid the neuromorphic simulation and enhance performance by directly inputting data into the network without prior conversion into spike signals. The initial layer of the network structure serves as the coding layer, employing a neuronal model that accepts both spike and non-spike signals. Essentially, neurons in the coding layer can process information from the original signal while maintaining sufficient accuracy to convert real-valued signals from non-spike data into spike sequences. This ensures that accuracy is primarily preserved with reduced reliance on simulation time. Since the parameters of the convolutional layer are adaptable, this coding approach is termed direct or adaptive coding. 

\subsubsection{\textbf{Learning Rules}}
Training of SNNs poses challenges due to the non-differentiable nature of spike events, unlike the successful backpropagation-based gradient descent learning in traditional ANNs \cite{javanshir2022advancements}. SNN training strategies include unsupervised learning, supervised learning and conversion from trained ANN models.

Unsupervised learning in SNNs adheres to the Hebbian rule, adapting synaptic connections based on received neuron data \cite{caporale2008spike}. The Spike-Timing-Dependent Plasticity (STDP) algorithm, an implementation of Hebb’s rule, captures changes in synapse efficacy relative to the timing of pre-synaptic and post-synaptic spikes. While STDP has been shown to approximate backpropagation update rules, it may not be ideal for training deep networks with high accuracy in deep learning applications \cite{dampfhoffer2023backpropagation}. Alternatively, backpropagation-based methods, effective in training deep networks, can directly train deep SNNs or indirectly convert pre-trained ANNs into their SNN equivalents.

SpikeProp \cite{bohte2002error} employs supervised learning, defining the cost function as the disparity between actual firing time and the target, and minimizes it using the gradient descent rule. Tempotron \cite{gutig2006tempotron}, another gradient descent training algorithm, utilizes the voltage disparity of the output neuron and the firing threshold as its cost function. While Tempotron achieves more efficient training compared to other algorithms, it's limited to binary classification problems. Initial research on ANN-to-SNN conversion, pioneered by \cite{perez2013mapping}, involved translating CNN units into biologically inspired spiking units with leaks and refractory periods. \cite{cao2015spiking} proposed a close connection between the transfer function of a spiking neuron and the activation of a Rectified Linear Unit (ReLU), which is now a standard model for ANNs. However, algorithms from \cite{diehl2015fast} and \cite{cao2015spiking} imposed significant limitations on the types of CNNs that could be converted. These limitations hindered the use of features commonly found in successful CNNs for visual classification, including max-pooling, softmax, batch normalization \cite{ioffe2015batch}, and even basic features of neuron models like biases. While significant progress has been made in the SNN domain, the majority of research efforts are concentrated on classification problems, resulting in limited literature addressing regression tasks.

\subsection{Visibility Enhancement}
The literature lacked solutions for spiking-based dehazing. In \cite{castagnetti2023spiden}, a directly trained SNN architecture utilizing surrogate gradient learning and backpropagation through time is introduced for Gaussian image denoising. In \cite{li2023deep}, a spiking-UNet architecture with multi-threshold spiking neurons is proposed for image segmentation and denoising. The authors employed a conversion and fine-tuning pipeline leveraging pre-trained U-Net models for training. The article also introduces a connection-wise normalization method to ensure accurate firing rates and adopts a flow-based training method for fine-tuning converted models, reducing time steps while maintaining performance. 

The Neural Architecture Search (NAS) proves effective in automatically selecting a neural network for single image dehazing. AutoDehaze \cite{mandal2022neural} was the first to utilize NAS with a search strategy based on gradients and hierarchical optimization at the network level for the purpose of image dehazing. The concept of NAS in spiking domain is introduced in \cite{kim2022neural} to develop a classification model that does not require training. This study utilizes NAS techniques to discover a model that effectively captures various patterns of spike activation across different datasets without the need for training. AutoSNN \cite{na2022autosnn} employs the NAS methodology to discover a power-efficient SNN model that outperforms manually designed SNNs in terms of accuracy and spike count. This is accomplished by directly training the super network and using a one-shot weight-sharing strategy that relies on an evolutionary algorithm. Efficient spiking-based dehazing architectures can be designed using these methods or combinations these methods.

\begin{figure*}[phbt]
\centering
\includegraphics[width=0.85\textwidth]{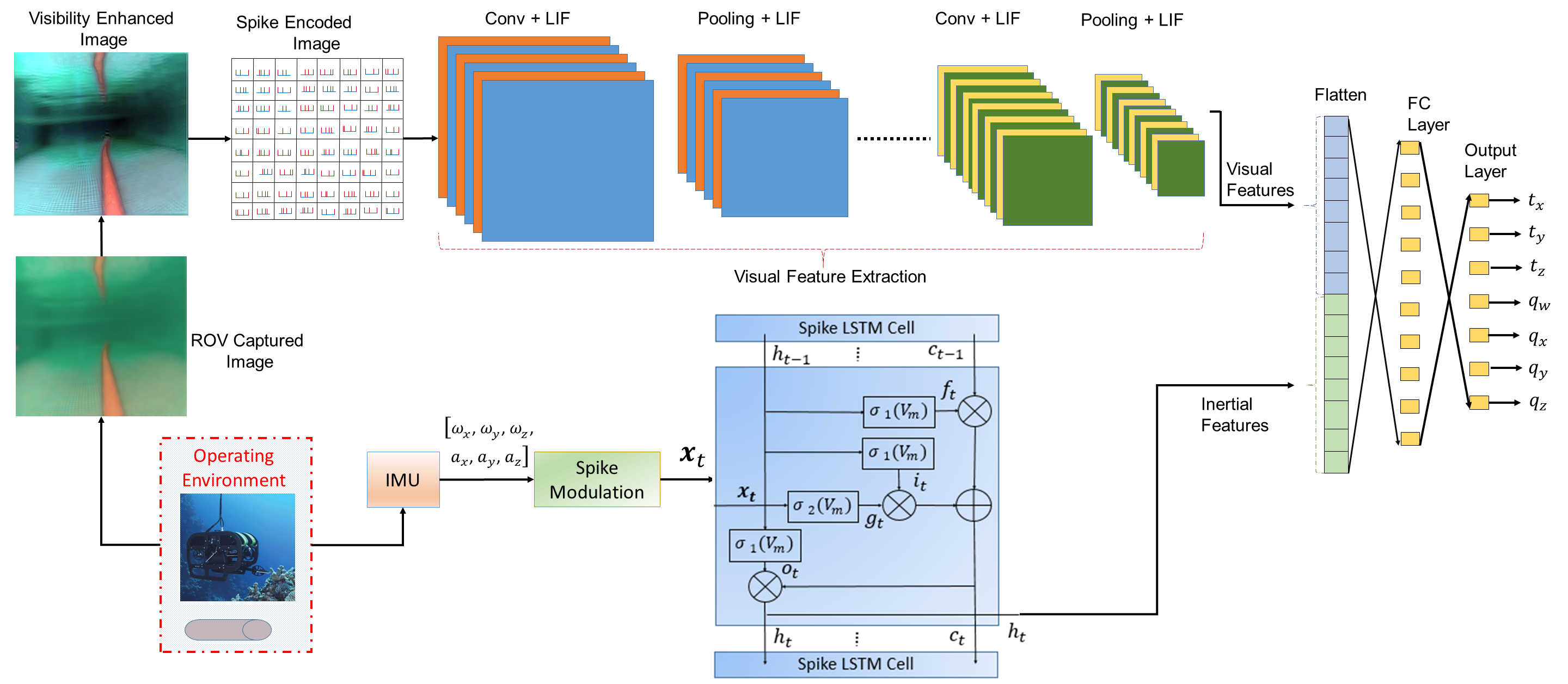}
\vspace{-0.3cm}
\caption{Schematic representation of spike-based CNN-LSTM hybrid framework}
\label{fig:actual_pose}
\end{figure*}

\subsection{Target Detection}
Spiking-YOLO architecture, trained for underwater scenarios, facilitates target detection in the spiking domain \cite{su2023deep}. The approach outlined in \cite{kim2020spiking} employs channel-wise normalization and signed neurons with imbalanced thresholds to address challenges like low firing rates in many neurons and inefficient leaky-ReLU implementation during DNN-SNN conversion. Channel-wise normalization adjusts weights based on the maximum achievable activation in each channel, eliminating extremely small activations and ensuring proper firing rates for accurate information transmission. The signed neuron with imbalanced threshold interprets both positive and negative activations and efficiently compensates for the leakage term in negative regions of leaky-ReLU. This method also preserves the discrete nature of spikes by introducing distinct threshold voltages for positive and negative regions, enabling the implementation of excitatory and inhibitory neurons, aligning with biological neuron principles.

\subsection{Target Pose Estimation}
Once the target structure is detected, its pose can be determined using an approach analogous to the one demonstrated in  \cite{chen2020g2l}. The output of the object detection network is fed into the translation-localization network for 3D segmentation and object translation prediction. Subsequently, the fine object point cloud, generated through predicted segmentation and translation, is transformed into a local canonical coordinate system. Within this system, a rotation localization network is trained to estimate the initial object rotation. In the spiking domain, the model can be realized either through CNN-SNN conversion or by directly designing and training the equivalent network.
\vspace{0.1cm}
\subsection{Vehicle Pose Estimation}
As there is a lack of existing literature on employing multi-modal pose estimation algorithms in the spiking domain, we are theoretically formulating a framework for this purpose. The translation and orientation vector of the actual ROV pose, \(_{}^{c}\textrm{p}_{a}=[t_x,t_y,t_z]\) and \(\theta u = f(q_w,q_x,q_y,q_z)\) is generated using a hybrid spike-based CNN-LSTM framework as shown in Fig.\ref{fig:actual_pose}.
\vspace{0.1cm}
\subsubsection{\textbf{Multi-layer Convolutional Spiking Neural Network}}
To recognize complex input patterns in high-dimensional spaces, multi-layer network architectures are essential, as they can learn hierarchical representations from input stimuli. In this context, a SNN model is utilized, featuring an input layer followed by intermediate hidden layers, aimed at extracting visual features. Input images are encoded as spike trains with a Poisson distribution, where the likelihood of spike generation correlates with pixel intensity. The hidden layers of the network serve to represent intermediate stages of feature hierarchies, incorporating spiking convolutional and pooling layers. At each time step, neurons convolve their input spikes with weight kernels to compute input currents, which are then integrated into their membrane potentials, denoted as \(V_{m}\). Neurons spike when \(V_{m}\) surpasses a threshold \(V_{th}\), resetting \(V_{m}\) to zero, or treat \(V_{m}\) as residual for the subsequent time step. The output of the visual and inertial feature extractors are combined to form a one-dimensional vector input for the fully connected spiking layers, which are followed by the output spiking layer of the regression head.
While spiking neural networks are commonly employed for classification tasks, where the output is determined by selecting the neuron with the highest firing rate \cite{chowdhury2022hardware}, this approach isn't suitable for regression tasks. Therefore, in this study, adjustments were made to the spiking neurons in the output layer to produce continuous values. Despite retaining synaptic currents and membrane potentials, these modified neurons no longer generate spikes. Consequently, the membrane potential \(V_{m}\) remains unaltered, and the model's output corresponds to the membrane potential of the output neuron. This adaptation yields a continuous and differentiable output, which is advantageous for regression tasks.
\vspace{0.1cm}
\subsubsection{\textbf{Spiking LSTM}}
The spiking LSTM unit consists of three interacting gates alongside their corresponding spike functions \cite{lotfi2020long}. The spike activations, denoted as \(\sigma _{1}\left ( V_{m} \right )\) and \(\sigma _{2}\left ( V_{m} \right )\), are applied to the respective neurons associated with each gate. These gate functions operate on the membrane potential \( V_{m}\) of neurons and produce either a spike or null output at each time step. The fundamental concept of spiking LSTM is similar to conventional LSTMs, revolves around the cell state \(c_{t}\), which governs the information flow between LSTM units. The forget gate \(f_{t}\) determines the information to be discarded, while the input gate \(i_{t}\) regulates the information entering the unit. An auxiliary layer, denoted as \(g_{t}\), is modulated by the spike activation \(\sigma _{2}\left ( V_{m} \right )\). Ultimately, the unit's output is determined based on the output gate \(o_{t}\) and the cell state \(c_{t}\). The distinct roles and characteristics of the various gates and states for an input spike train \(\left \{ \mathbf{x}_1,\mathbf{x}_2,...,\mathbf{x}_n \right \}\) are elaborated as follows:
\vspace{-0.1cm}
\begin{align}
\begin{split}
    f_t=\sigma_1(w_{h,f}h_{t-1}+w_{x,f}x_{t}+b_{h,f}+b_{x,f})\\
    i_t=\sigma_1(w_{h,i}h_{t-1}+w_{x,i}x_{t}+b_{h,i}+b_{x,i})\\
    g_t=\sigma_2(w_{h,g}h_{t-1}+w_{x,g}x_{t}+b_{h,g}+b_{x,g})\\
    o_t=\sigma_1(w_{h,o}h_{t-1}+w_{x,o}x_{t}+b_{h,o}+b_{x,o})\\
    c_t=f_t\bigodot c_{t-1}+i_t\bigodot g_{t}\\
    h_t=o_t\bigodot c_t
    \label{equation:spike_LSTM_states}
\end{split}
\end{align}
where \(\bigodot\) represents the Hadamard product. 

\section{Discussion} \label{sec:discussion}
In this research, we introduced an innovative approach involving neuromorphic methodologies to enhance task-oriented applications within underwater robotics. Reviewing the literature reveals that SNN based architectures have the potential to reduce computational demands and power consumption compared to conventional learning-based methods. Our proposed unified framework demonstrates the versatility of neuromorphic concepts across various stages, encompassing perception, pose estimation, and haptic-guided conditional control of underwater vehicles. Leveraging the advantages of SNNs over traditional learning-based frameworks, the integration of SNN modules offers the prospect of reduced power consumption while maintaining comparable performance to conventional models. We assume that this conceptual framework holds promise for triggering a transformative shift in underwater robotics, fostering greater efficiency and autonomy while concurrently addressing energy consumption concerns.

However, it's imperative to acknowledge certain limitations. These include the limited literature addressing the effectiveness of utilizing membrane potential for training SNN models, the necessity for selecting appropriate neuron models that can yield optimal performance across diverse applications, and the development of spiking-based regression frameworks tailored for complex user-defined tasks. Additionally, the absence of sufficient data suitable for training SNN models with corresponding ground truth for pose estimation in the underwater domain presents a challenge. The implementation of each neuromorphic module for perception, multi-modal pose estimation, and object detection will uncover practical implementation challenges and pave the way for future research endeavors. Our discussion delves into the implications and potential challenges associated with adopting neuromorphic methodologies in underwater robotics, while also illuminating avenues for future research and development in this promising field.

\section{Conclusions and Future Research  \label{sec:conclusions}}
This research marks a significant milestone in the advancement of underwater robotics through the innovative integration of neuromorphic methodologies. By harnessing the potential of spiking neural network (SNN) architectures, the proposed framework offers a compelling solution to the challenges faced by conventional learning-based methods. The versatility of neuromorphic concepts demonstrated across various stages, from perception to pose estimation and haptic-guided control, highlights the transformative potential of this approach in enhancing efficiency and autonomy while addressing energy consumption concerns.

Moving forward, future research endeavors should focus on addressing several key challenges and exploring opportunities for further refinement and development. Firstly, the effectiveness of utilizing membrane potential for training SNN models warrants further investigation, as does the selection of appropriate neuron models to optimize performance across diverse applications. Additionally, the development of spiking-based regression frameworks tailored for complex user-defined tasks presents an area ripe for exploration.

Furthermore, the limited availability of suitable data for training SNN models, particularly in the context of pose estimation in the underwater domain, poses a significant challenge. Addressing this limitation and uncovering practical implementation challenges associated with each neuromorphic module will be crucial for advancing the field. Despite these challenges, the integration of neuromorphic technologies holds immense promise for revolutionizing underwater robotics and advancing our exploration of marine environments. Through continued research and development efforts, we can further unlock the full potential of neuromorphic approaches in underwater robotics, paving the way for exciting new discoveries and applications in the future.

\section*{Acknowledgment}
The authors acknowledge the project “Heterogeneous Swarm of Underwater Autonomous Vehicles”, a collaborative research project between the Technology Innovation Institute (Abu Dhabi)  and Khalifa University  (contract no. TII/ARRC/2047/2020). This work is also supported by Khalifa University under Awards No. RC1-2018-KUCARS-8474000136. 


\end{document}